\documentclass[journal]{IEEEtran}

\ifCLASSINFOpdf
\else
\fi

\usepackage{amsmath}
\usepackage{amsfonts}

\usepackage{nomencl}
\makenomenclature

\usepackage[ruled,vlined]{algorithm2e}

\usepackage[table]{xcolor}
\usepackage{makecell}
\usepackage{multirow}

\usepackage{graphicx}
\usepackage{caption}
\usepackage{subcaption}
\usepackage{svg}

\usepackage[
backend=biber,
style=ieee,
sorting=none
]{biblatex}
\usepackage{hyperref}

\begin{document}

\title{pmuBAGE: The Benchmarking Assortment of Generated PMU Data for Power System Events - Part I: Overview and Results\\
}

\author{Brandon~Foggo,~\IEEEmembership{Member,~IEEE,}
        Koji Yamashita,~\IEEEmembership{Member,~IEEE,}
        Nanpeng~Yu,~\IEEEmembership{Senior Member,~IEEE,}
\thanks{B. Foggo, K. Yamashita, and N. Yu are with the Department
of Electrical and Computer Engineering, University of California, Riverside, CA 92501 USA. E-mail: nyu@ece.ucr.edu.}
}

\newcommand{\dee}[2]{\frac{\partial{#1}}{\partial{#2}}}

\maketitle

\begin{abstract}

We present pmuGE (phasor measurement unit Generator of Events), one of the first data-driven generative model for power system event data. We have trained this model on thousands of actual events and created a dataset denoted pmuBAGE (the Benchmarking Assortment of Generated PMU Events). The dataset consists of almost 1000 instances of labeled event data to encourage benchmark evaluations on phasor measurement unit (PMU) data analytics. The dataset is available online for use by any researcher or practitioner in the field. PMU data are challenging to obtain, especially those covering event periods. Nevertheless, power system problems have recently seen phenomenal advancements via data-driven machine learning solutions - solutions created by researchers who were fortunate enough to obtain such PMU data. A highly accessible standard benchmarking dataset would enable a drastic acceleration of the development of successful machine learning techniques in this field. We propose a novel learning method based on the Event Participation Decomposition of Power System Events, which makes it possible to learn a generative model of PMU data during system anomalies. The model can create highly realistic event data without compromising the differential privacy of the PMUs used to train it. The dataset is available online for any researcher to use at the pmuBAGE \href{https://github.com/NanpengYu/pmuBAGE}{Github Repository}.

Part I - This is part I of a two part paper. In part I, we describe a high level overview of pmuBAGE, its creation, and the experiments used to test it. Part II will discuss the exact models used in its generation in far more detail.

\end{abstract}


\begin{figure}[t]
    \centering
    \begin{subfigure}{0.49\linewidth}
        \centering
        \textbf{pmuBAGE}\par\medskip
        \includegraphics[width=\linewidth]{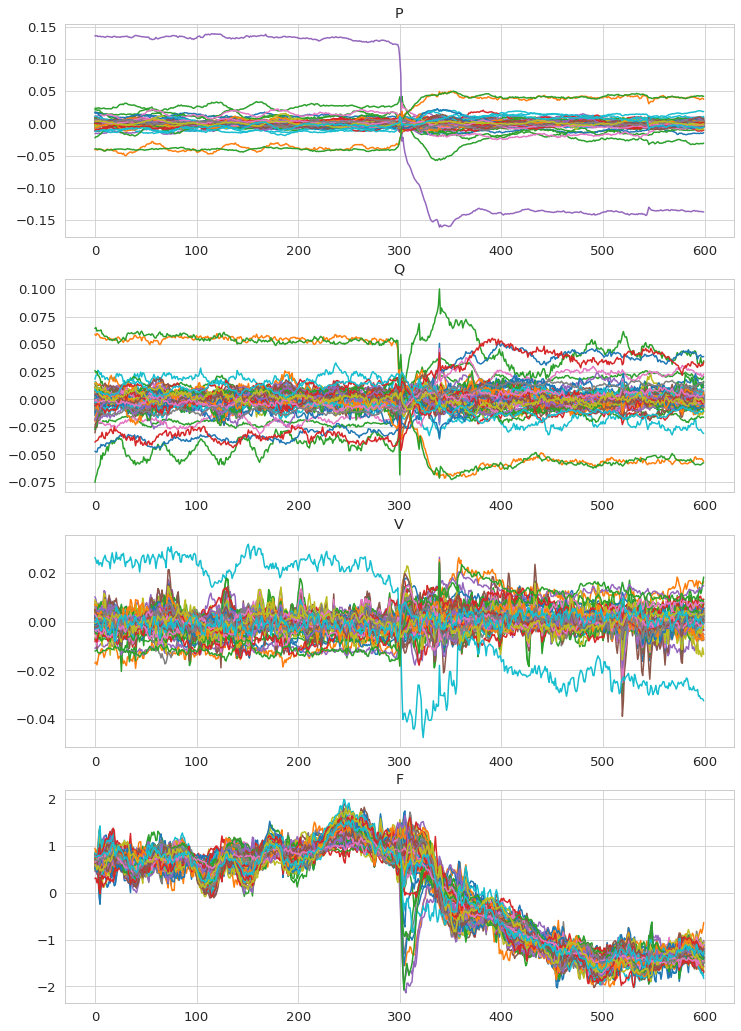}
    \end{subfigure}
    \begin{subfigure}{0.49\linewidth}
        \centering
        \textbf{Actual Data}\par\medskip
        \includegraphics[width=\linewidth]{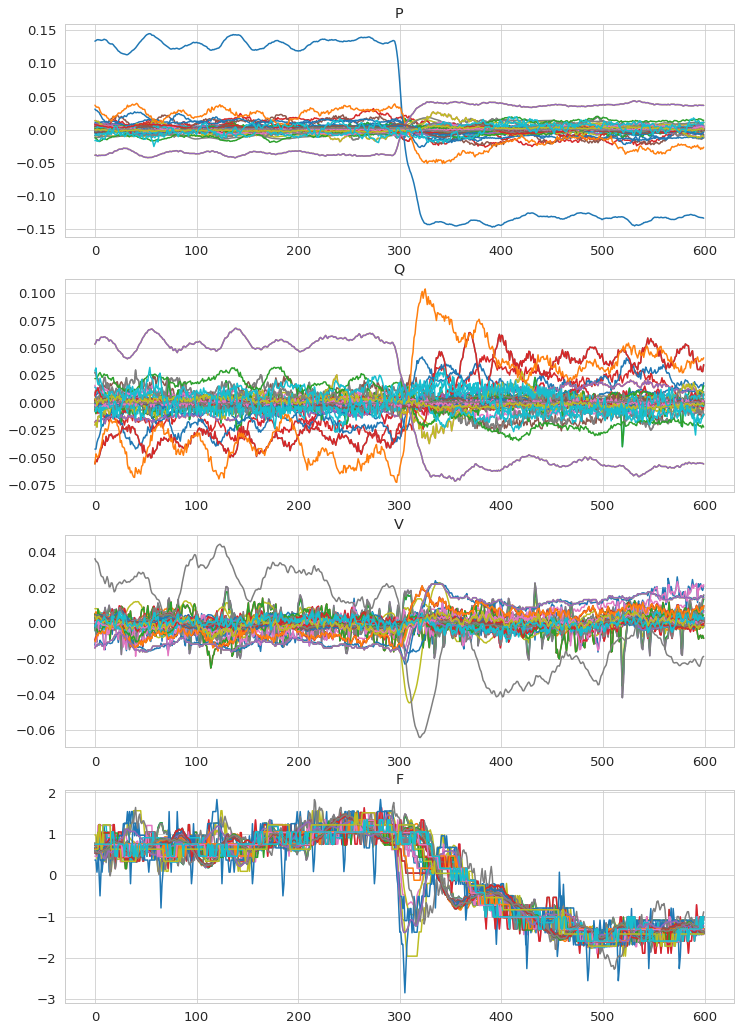}
    \end{subfigure}
    
    \caption{(Left) A generated frequency event in pmuBAGE. (Right) A real frequency event of similar type. Different colors represent different PMUs (100 signals). The interval between two time indices is 1 / 30 seconds. The full window corresponds to $20$ seconds of data. The presented data is scaled to per unit values and each timeseries is shifted to have a mean of zero. The four sub-figures represent real power, reactive power, voltage magnitude, and frequency from top to bottom.} 
    \label{fig:fevent_display}
\end{figure}

\begin{figure}[t]
    \centering
    \begin{subfigure}{0.49\linewidth}
        \centering
        \textbf{pmuBAGE}\par\medskip
        \includegraphics[width=\linewidth]{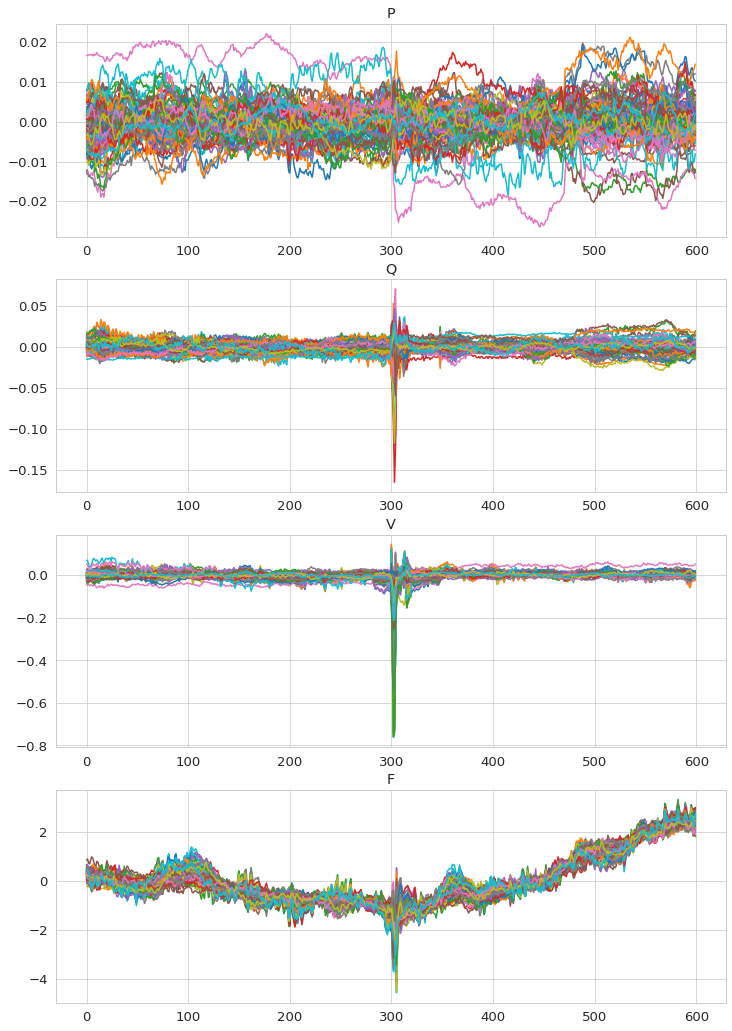}
    \end{subfigure}
    \begin{subfigure}{0.49\linewidth}
        \centering
        \textbf{Actual Data}\par\medskip
        \includegraphics[width=\linewidth]{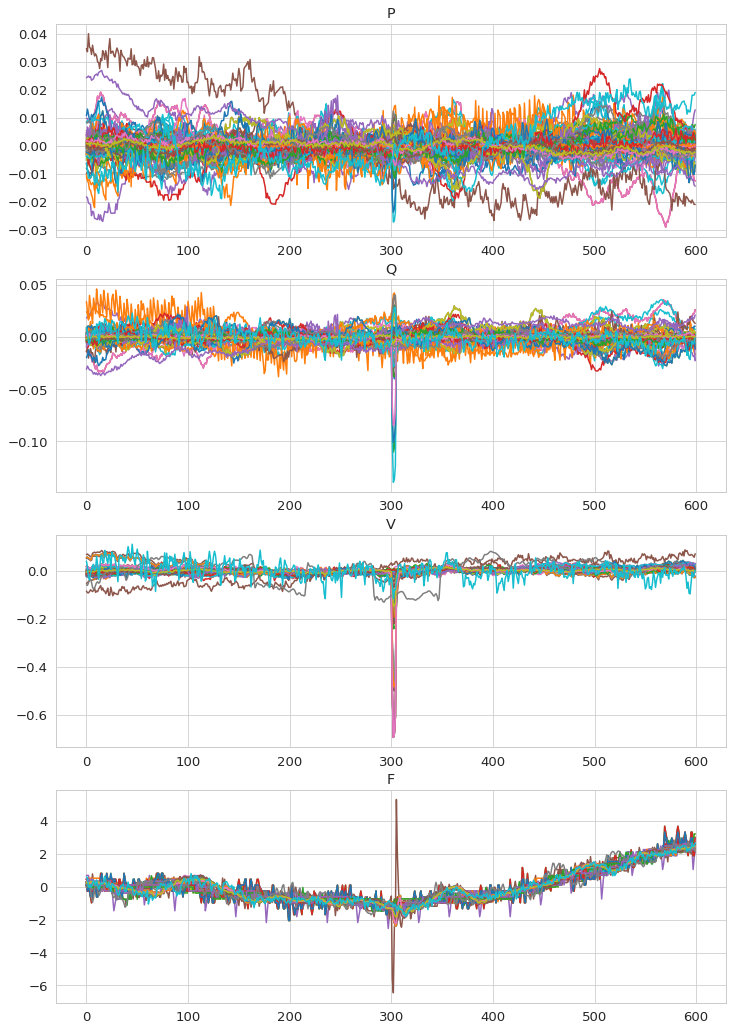}
    \end{subfigure}
    
    \caption{(Left) A generated voltage event in pmuBAGE. (Right) A real frequency event of similar type. Different colors represent different PMUs (100 signals). The interval between two time indices is 1 / 30 seconds. The full window corresponds to $20$ seconds of data. The presented data is scaled to per unit values and each timeseries is shifted to have a mean of zero. The four sub-figures represent real power, reactive power, voltage magnitude, and frequency from top to bottom.} 
    \label{fig:vevent_display}
\end{figure}

%
\IEEEpeerreviewmaketitle

\section{Introduction}
\IEEEPARstart{S}{olutions} to any power system dynamic study require realistic dynamic response data. 
Obtaining realistic phasor measurement unit (PMU) data that are used by the transmission system operator (TSO), regional system operator (RTO) or electric utilities has always been a bottleneck in academia. 
Large-scale IEEE dynamic test cases, such as the 50-generator (145-bus) system \cite{IEEE50}, have been widely leveraged to cope with this. 
However, generally accepted dynamic models, such as combined cycle gas turbine models, and nontraditional power components, such as renewable energy sources (RES), are missing in such dynamic test cases. 
Although generic dynamic models for turbine-governors \cite{PES_TR1} have been publicly available and implemented into many commercially available dynamic simulation software, the widely accepted model parameters are not fully provided.

The \textit{power system engineering research center} (PSERC) has begun to create synthetic dynamical data by simulating more realistic North American Eastern and Western Interconnection models \cite{PSERC2021}. 
Many non-data driven modelling approaches to create synthetic data can be found in literature. Much of this research builds on each other in a modular way. For example, reference \cite{7515182} uses geographic properties to automatically generate synthetic networks, and reference \cite{7459256} focuses on the placement of generator substations and assignment of transmission line parameters to an existing topology. Reference \cite{8334287} automatically adjusts the parameters of the placed generators according to fuel type and the statistics of real generators in the same geographic region. References \cite{8312059} and \cite{9174809} focus on improving load properties in synthesized grids. Reference \cite{8333771} automatically adds realistically placed voltage controllers onto an existing synthetic grid. Finally, reference \cite{9216129} focuses on integrating synthetic transmission systems with synthetic distribution systems. References \cite{7725528} and \cite{en10081233} describe statistical properties that can be checked to ensure that a synthetic grid is realistic to its geographical region.

The results of these attempts surely help demonstrate some particular dynamic aspects of power system events more accurately - such as poorly damped natural/forced oscillations and frequency drop following disturbances. 
However, we have no all-encompassing or one-size-fits-all dynamic model. 
Besides, the parameterization of the aforementioned generic model, specifically RES models, becomes more difficult mainly due to its accelerating spatiotemporal dispersion, which results in insufficient simulation accuracy (i.e., less credible simulation results), especially for the future grid analysis with high penetration of the RES. 
Superimposing realistic noise to the simulated response is also not trivial. 
Therefore, real-world measurement data is craved by both academics and industry engineers.

However, access to real-world PMU data is quite limited. Although some TSO/RTOs have started to provide sample PMU data to researcher, most do not offer to serve them. The primary factor of hesitation for the release of this data is the potential risk of exposing any vulnerabilities in the bulk power system. Most of the time, power system research teams will go through several steps to setup small collaborations with the utility companies who own this PMU data. This will result in a small amount of data spanning a somewhat random time-span, and will be accessible by that research team for about the length of time that it requires to perform one or two innovative research projects.

There are two meta-level effects to this process. The first effect is that initiating a project on actual data in this field has a considerable time overhead, limiting the amount of quality research that comes out. The second effect is significant heterogeneity of results on the same problem. Suppose, for example, that two research teams developed a data-driven algorithm with data from the same geographical location for the same problem. However, the first team used spring data, and the second used summer data. The first team's algorithm then reports a $2\%$ improvement in performance over the second team. However, summer data is more unwieldy than spring data. Thus, some improvement must have come from using a more well-behaved dataset. In no way can we reliably determine the size of this contribution, so we are left in a scenario in which we do not honestly know which algorithm has better performance.

Realistic \textit{generated} data, on the other hand, is more readily accepted for release by TSO/RTOs because such data is no longer treated as real. By generated data, we refer specifically to the process of creating synthetic data by sending random noise through a deep neural network trained on real data. Such deep generative models promise to boost power system dynamic studies, specifically for event detection and classification with big data analysis.

The dataset presented in this paper, pmuBAGE, will bring back the advantages of using standard IEEE dynamic test cases - i.e., the accessibility and homogeneity of results - while maintaining the realism and difficulty of dealing with real PMU data. We encourage any researcher in this field to provide benchmarks of their algorithms using this dataset. Two example events are displayed in Figures \ref{fig:fevent_display} and \ref{fig:vevent_display}.

pmuBAGE is the result of training a novel generative model, called the pmuGE (the phasor measurement unit Generator of Events), and training it on over $1000$ real events gathered over two years on the bulk U.S. Power Grid. While pmuBAGE is an immediately available dataset, the generative model will be described in excruciating detail in the following subsections of this two parts paper. Thus, if some researchers desired to create an updated version of pmuBAGE using new PMU data, the instructions for doing so are readily available in this paper. The model preserves the privacy of the PMUs used to train it. 
 
We'd like to note a few preliminary works of this type in literature. In these works, performed in references \cite{zheng2019creation, 8784681, 9361704}, a simple Generative Adversarial Network (GAN) architecture was trained on data simulated via IEEE dynamic test cases. These works are important in showing the feasibility of training generative models on PMU data during events. However, the limitations of the data used to train these models - that the corresponding training datasets do not come from real synchrophasor data - resulted in a synthetic datasets with unrealistic noise and PMU-specific behaviors.

To our knowledge, the present work is the first to attempt a fully data driven approach to generating synthetic synchrophasor data using large-scale real world PMU data.

The contributions of this paper are as follows:
\begin{itemize}
    \item By decomposing the PMU data into a static statistical component and a dynamic physical component, separating dynamic components into inter-event components and intra-event components, and using state-of-the-art probabilistic programming methods and deep cascaded convolutional generative models, we are able to create much more realistic synthetic event data than was previously possible.
    \item We introduced a new QR-reorthogonalization trick to reintegrate the inter-event signatures with intra event signatures and sparse signatures, which are not typically statistically independent. This significantly increases the ease of training our model.
    \item By introducing several feature matching training loss functions to our model, including a completely new loss function which we denote the ``quantile loss", we are able to capture the distribution of the aforementioned statistical component extremely tightly.
    \item We trained our model on the largest real-world synchrophasor dataset to date. This has allowed us to capture the PMU-varying and time-varying dynamics of power system event data more accurately.
    \item We provide the first comprehensive set of realistic synthetic events covering many types of event causes, including previously under-modelled event types such as those involving lightning strikes and renewable generation.
\end{itemize}

Example applications of pmuBAGE include experimenting/benchmarking for power system event detection \cite{9648034, xie2014dimensionality, brahma2016real, negi2017event, liu2019data, ge2015power, li2017fast, wang2020frequency, zhou2018ensemble, shi2020online, cui2018novel}, event classification \cite{9431702,hannon2019real,dahal2012preliminary,li2019unsupervised,li2019hybrid,nguyen2015smart}, and missing value replacement (especially during events) \cite{de2018evaluation, chatterjee2019robust, liao2018estimate, osipov2020pmu, konstantinopoulos2020synchrophasor, gao2015missing, huang2016data, 9468345}

Part I - This is part I of a two part paper. In part I, we describe a high level overview of pmuBAGE, its creation, and the experiments used to test it. Part II will discuss the exact models used in its generation in far more detail.

pmuBAGE is available online in our \href{https://github.com/NanpengYu/pmuBAGE}{Github Repository}.
 
 \section{Overall Framework}
 
 \begin{figure}
     \centering
     \includegraphics[width=\linewidth]{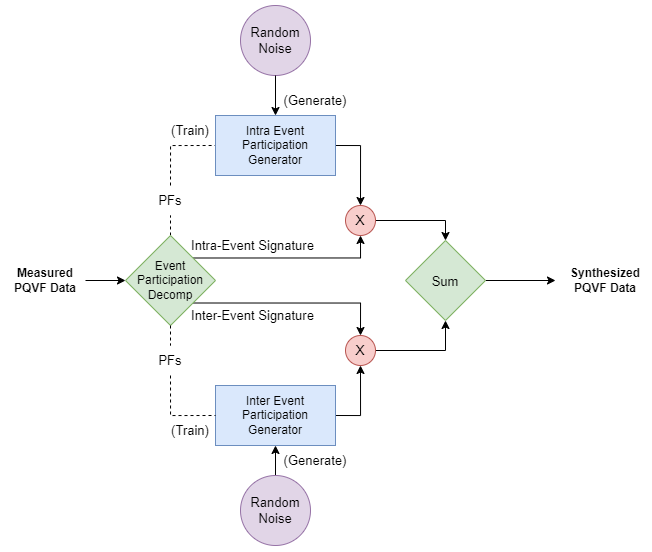}
     \caption{The High Level Framework of our model.}
     \label{fig:overveiw}
 \end{figure}
 
Our goal is to create a realistic PMU dataset for power system events without compromising the privacy of the PMUs used to train it. The ``Event-Participation" (EP) decomposition transformation is our primary device for achieving this. This decomposition separates event tensors into Event Signatures, shared across all PMUs, and participation factors specific to each PMU \cite{9468345}. By using the Event Participation Decomposition, the data properties that \textit{could} compromise PMUs are extracted into the participation factors, while the parts that cannot are extracted into event signatures. This allows us to maintain the properties of our dataset that do not vary with PMUs (i.e., the event signatures) in their exact form. We then need only perform statistical/generative modeling on the participation factors of this decomposition.
 
A very high-level view of our framework is provided in Figure \ref{fig:overveiw}. Standardized PQVF (real power, reactive power, voltage magnitude, frequency) data tensors are first fed into a module denoted ``EP" (short for Event-Participation Decomposition). The EP module decomposes each tensor into four building blocks - inter-event signatures, intra-event signatures, inter-event participation factors, and intra-event participation factors. No generative modeling is required for either of the event-signature components. On the other hand, the Participation factor components cannot be used directly since they carry all of the PMU specific information.

The synthetic Intra-Event components are created via a deep generative probabilistic program resembling a Generative Adversarial Network (GAN). The inter-event participation factors, being much less nuanced in their distributions, are modelled via a statistical simulation.
 
\section{The Event-Participation Decomposition}
\begin{figure*}[t]
    \centering
    \begin{subfigure}{0.3\linewidth}
        \includegraphics[width=\linewidth]{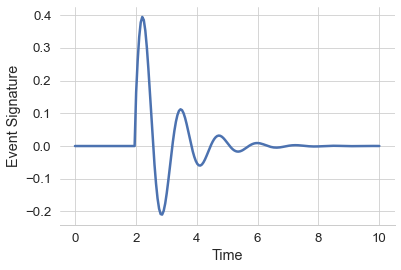}
    \end{subfigure}
    \begin{subfigure}{0.3\linewidth}
        \includegraphics[width=\linewidth]{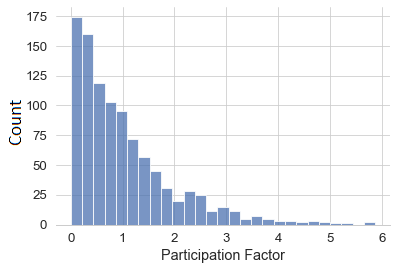}
    \end{subfigure}
    \begin{subfigure}{0.3\linewidth}
        \includegraphics[width=\linewidth]{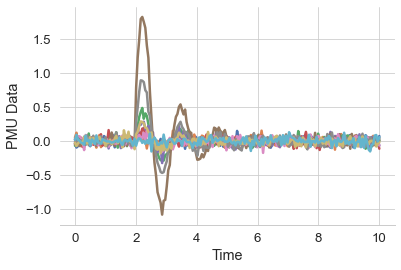}
    \end{subfigure}
    \caption{(Left) An event signature. (Middle) Corresponding Participation Factor distribution. (Right) Resulting PMU data array (with some white noise added for realism).}
    \label{fig:EPConcept}
\end{figure*}

Our generative model uses the Event-Participation (EP) decomposition of PMU event data. The decomposition was originally used in reference \cite{9468345} wherein its time-independent component was leveraged for missing value replacement. Event-Participation decomposes events into several dynamic components, called event signatures, and a corresponding set of static components (which vary over PMUs), called participation factors. Each event signature captures a distinct component of the overall shape of the event, and the corresponding participation factor captures how much each PMU undergoes that behavior. This idea has its roots in the small-signal stability analysis of power systems. A simple conceptual example of this decomposition is shown in Figure \ref{fig:EPConcept}.

Before we describe how to compute this decomposition, we will first give some motivation to its use in the present work. The decomposition makes the conceptualization of a generative model much simpler. We can think of each participation factor as a set of samples from an independent identically distributed distribution that depends on the event signature. We can thus, in turn, think of each event-participation factor as a composition of two maps - the first of which takes the event signature and returns a probability distribution, and the second of which takes the distribution and returns samples. Since no learning is required to imitate a sampling mechanism, we need only focus on the former of these maps. 

To put this into mathematical terms, suppose we are given an event signature $e \in \mathbb{R}^{T}$ \nomenclature{$e$}{Event Signature} (where $T$ is the temporal length of each instance of data\nomenclature{$T$}{Temporal length of each instance of data}). Then we need only find a probability distribution $\mu_{\theta}(e)$ for that event (where $\theta$ denotes any parameters used to specify that probability distribution). Once $\mu_{\theta}(e)$ is learned (as a full function of $e$), we can take any event signature - whether it exists in our training dataset or not - and create a set of PMU data adhering to that signature by sampling from it. \nomenclature{$\mu_{\theta}(e)$}{Participation Factor Probability Distribution}

Note that we need not be limited to a single event signature and participation factor. It is, in fact, far more realistic to assume that multiple dynamical modes (i.e., event signatures) coincide - each with their own set of participation factors.

Most applications using this decomposition require online use and often deal with data quality problems. As a result, those applications must use advanced techniques to compute this decomposition. However, the present work does not have such limitations. At the end of the day, when given a full tensor of clean PMU data, event-participation decomposition is nothing more than a regular tensor decomposition - just with some interpretation added to the factors.

\section{Numerical Results}

\subsection{Training Dataset}

Our training dataset consists of $PQVF$ data for $620$ Voltage Events and $84$ Frequency Events. Data was collected by Pacific Northwest National Laboratory from $180$ simultaneous Synchrophasor measurements across the Eastern Interconnection of the U.S. Transmission Grid at a sampling rate of $30 Hz$. The original Synchrophasor data consisted of Voltage Magnitude, Voltage Angle, Current Magnitude, Current Angle, and Frequency decomposed as symmetrical components (as opposed to $A$, $B$, $C$ phase data). Some Synchrophasors reported all three symmetric components for each data type (Positive Sequence, Negative Sequence, and Zero Sequence), but most only reported the Positive Sequence component. Thus only Positive Sequence values were used to train our model. $PQVF$ data was derived from the values mentioned above.

Each event in our training dataset consists of $20$ seconds ($600$ samples) of data, with the event's start placed at the $10$ second mark ($300$-th sample). The exact event timing was not always reported accurately, so expert analysis either affirmed or adjusted these start times by a few fractions of a second. Frequent missing data samples appeared during event windows. PMUs that exhibited such values were removed from the dataset, but only for the event where they had missing data. Thus the PMUs used in training for one event are not necessarily the same as the set of PMUs used to train another. 

\subsection{Training Details}
\begin{figure}[]
    \centering
    \begin{subfigure}{0.49\linewidth}
        \includegraphics[width=\linewidth]{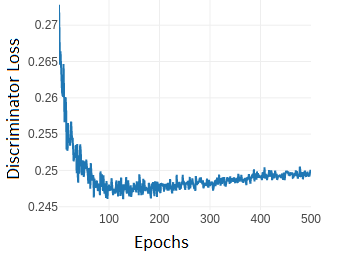}
    \end{subfigure}
    \begin{subfigure}{0.49\linewidth}
        \includegraphics[width=\linewidth]{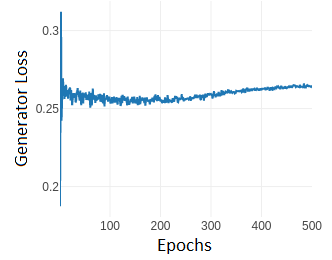}
    \end{subfigure}
    \caption{(Left) Discriminator Loss Curve over Epochs. (Right) Generator Loss Curve over Epochs.}
    \label{fig:training}
\end{figure}

The parameters of our model were trained for $500$ epochs with a batch size of $50$ events per batch. The Adam optimizer was employed, and an $l_2$ weight regularization term with a coefficient of $0.25$ was applied. Generative Adversarial Networks are known to have notorious training difficulties. The most common difficulty occurs when the discriminator converges too quickly. This forces the discriminator's feedback to the generator to have vanishing gradients. To avoid this, the learning rates set for the discriminator's parameters were significantly lower than that of all other parameters. Expressly, we set the learning rate of discriminator parameters to $1\times10^{-5}$ while all other parameters used a learning rate of $1\times 10^{-3}$.

Training this model proceeds relatively smoothly. Training curves are shown in Figure \ref{fig:training}. Both loss functions dip to low values quickly during training, followed by a somewhat slow-moving period in which both curves converge to their near-optimal values. Here we have plotted the GAN loss used in training, not the slew of feature matching losses described earlier. Loss curves for those feature matching losses are similarly stable but converge much more quickly.

\subsection{Similarity Analysis}
While we want the resulting generated events to be similar to the training set, we certainly do not want them to be identical. If identicality were our goal, then there would have been no problem to solve in the first place - we would have just copied and pasted the dataset.

To ensure some level of non-similarity to the original dataset, we performed a correlation analysis. To do so, we partitioned the datasets in two different ways - first, by Event ID, and second, by PMU ID. For the Event ID partition, we went through each event and computed the full tensor inner product of the synthetic event tensor and the measured event tensor (recall that the means and standard deviations of these datasets are already equal to zero and one respectively). For the PMU ID partition, we went through each PMU and calculated the full tensor inner product for that PMU \textit{after concatenating all events that the PMU is in}.

The Event ID partition will help us identify if any of the synthetic events in our dataset are too similar to the corresponding measured event (the event with the same set of event signatures as the synthesized tensor). The PMU ID partition will help us identify if any PMUs have been compromised by the synthetic dataset. 

\begin{figure}[]
    \centering
    \begin{subfigure}{0.49\linewidth}
        \includegraphics[width=\linewidth]{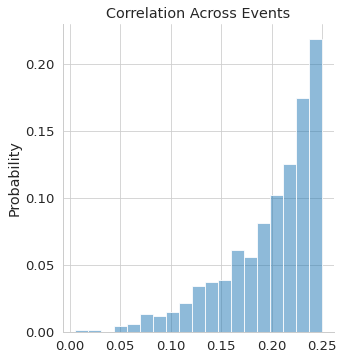}
    \end{subfigure}
    \begin{subfigure}{0.49\linewidth}
        \includegraphics[width=\linewidth]{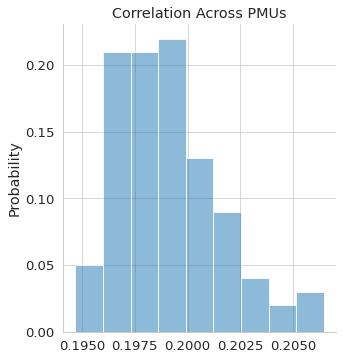}
    \end{subfigure}
    \caption{Correlation histograms between the synthetic and measured PMU datasets partitioned by (Left) Event ID (Right) PMU ID.}
    \label{fig:corr}
\end{figure}

Two histograms of correlation values are plotted in Figure \ref{fig:corr}. On the left, each datapoint used in calculating the histogram was an Event ID partitioned correlation value. There is a peak at correlation $0.25$, but this is the maximum of such correlation. As such, we deem that no events are too similar to their original incarnation. On the right of Figure \ref{fig:corr}, each datapoint used in calculating the histogram was a PMU ID partitioned correlation value. No correlation exceeds $0.21$, so we deem that no PMU has been compromised.

\subsection{Inception-Like Scoring}
One of the most well-accepted ways to analyze the quality of generated data samples is through a measure called the ``Inception Score." A universally agreed-upon third-party classification model, called the Inception Model, is trained on a set of measured data to calculate an inception score. Generated samples are then fed into the Inception Model, and the classification accuracy of the Inception Model on those generated images is reported. A high classification accuracy indicates that the generated images are of high quality.

Unfortunately, we can use no standard Inception Model for our generated Event Samples. This is not surprising since, to our knowledge, this paper represents the first attempt to create such data. We will thus assess the quality of these event samples with an approximation of the Inception Score. To do this, we took a relatively standard ResNext \cite{Xie2016}  Model (equivalent to each of the ``Feature Extraction Maps" presented above) and trained it to classify event types with class labels ``Frequency", and ``Voltage". Training of this classification model was performed over $200$ epochs with a batch size of $50$ using the Binary Cross Entropy loss function. Training was performed with the Adam optimizer with a learning rate of $1\times 10^{-3}$. Training proceeded smoothly with near monotonically-decreasing loss.

Training was performed twice - once with the synthetic data and once with the measured data. Each training data point was 20 second event window - either synthetic or real depending on the experiment - unmodified from its original form. We then tested each trained model on the synthetic and measured PMU data. This gives us four different train-test scenarios:
\begin{enumerate}
    \item Train on Synthetic, test on Synthetic
    \item Train on Synthetic, test on Measured
    \item Train on Measured, test on Synthetic
    \item Train on Measured, test on Measured
\end{enumerate}

We are interested in the difference between the Synthetic-Synthetic, Measured-Measured scores and the Synthetic-Measured, Measured-Synthetic Scores. Suppose the cross scores (Synthetic-Measured, Measured-Synthetic) are not significantly degraded from the self scores (Synthetic-Synthetic, Measured-Measured). In that case, we deem the quality of the generated samples to be high.

In terms of interpretation of these scores, the Synthetic-Measured score indicates how well a model trained on these generated events will perform in practice. In contrast, the Measured-Synthetic score indicates how well a model trained on measured data would benchmark against pmuBAGE.

Results are reported in Table \ref{tab:inception}. Since our measured and synthetic datasets are imbalanced (with a ratio of about ${7:1}$ in favor of voltage events), we report F1 and F2 scores alongside classification accuracy. In training each classification scenario, we multiplied each training loss for voltage events by $1/7$ so that their back-propagation gradient descents would be weighted equally to those of frequency events.

\begin{table}[h]
    \centering
    \def\arraystretch{1.5}
    \begin{tabular}{l | l  l  l}
    \hline
         Training-Testing & Acc. & F1 & F2 \\ \hline
         Synthetic-Synthetic & $99.9\%$ & $94.3\%$ & $93.3\%$ \\
         \textbf{Synthetic-Measured} & $\mathbf{94.3\%}$ & $\mathbf{94.2\%}$ & $\mathbf{92.8\%}$ \\
         Measured-Measured  & $99.8\%$& $94.4\%$ &  $91.2\%$ \\
         \textbf{Measured-Synthetic} & $\mathbf{93.2\%}$ &  $\mathbf{94.3\%}$ & $\mathbf{92.7\%}$ \\ \hline
    \end{tabular}
    \caption{Self Scores and Cross Scores across different train-test scenarios. Accuracy, F1 scores, and F2 scores are reported.}
    \label{tab:inception}
\end{table}

There appears to be no significant degradation in $F1$ or $F2$ scores in cross-comparisons compared to self comparisons, but there is a notable drop in accuracy. However, this drop is less than $7\%$, and all values are still greater than $90\%$. However, since the dataset is very imbalanced for this problem, the $F$ scores are more relevant since the accuracy measure is far less sensitive to the correct identification of frequency events than they are to voltage events (even if we misclassified every frequency event, we would still obtain an accuracy of $85\%$).

\subsection{Synthetic Event Images}
\begin{figure*}[t]
    \centering
    
    \begin{subfigure}{0.33\linewidth}
        \includegraphics[width=\linewidth]{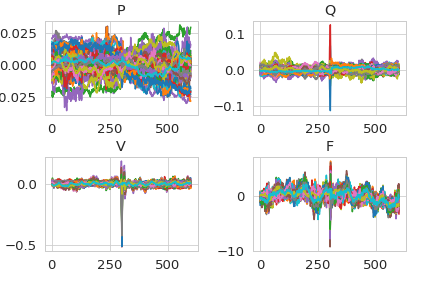}
    \end{subfigure}\hfill
    \begin{subfigure}{0.33\linewidth}
        \includegraphics[width=\linewidth]{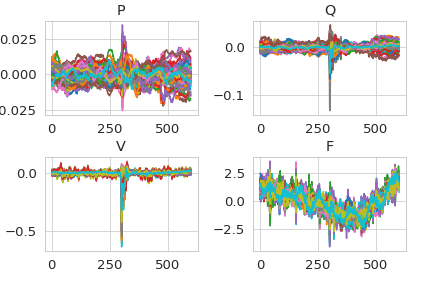}
    \end{subfigure}\hfill
    \begin{subfigure}{0.33\linewidth}
        \includegraphics[width=\linewidth]{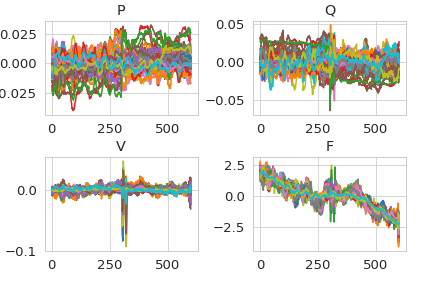}
    \end{subfigure}\hfill
    
    \caption{Samples of generated full event tensors designated as ``voltage events" caused by lightning strikes. All time stamps on the horizontal axis correspond to $1/30$ seconds, and the full event window comprises $20$ seconds of synthetic event data. The presented data is scaled to per unit values and each timeseries is shifted to have a mean of zero.}
    \label{fig:vgensamples_lightning}
\end{figure*}

\begin{figure*}[t]
    \centering
    
    \begin{subfigure}{0.33\linewidth}
        \includegraphics[width=\linewidth]{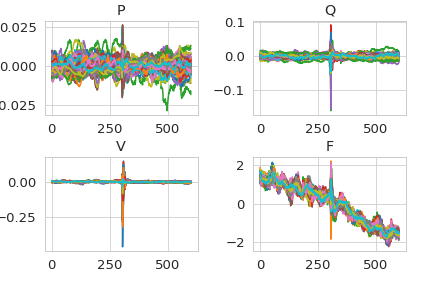}
    \end{subfigure}\hfill
    \begin{subfigure}{0.33\linewidth}
        \includegraphics[width=\linewidth]{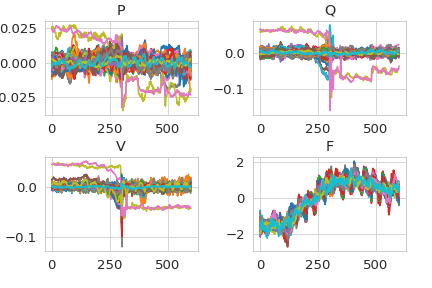}
    \end{subfigure}\hfill
    \begin{subfigure}{0.33\linewidth}
        \includegraphics[width=\linewidth]{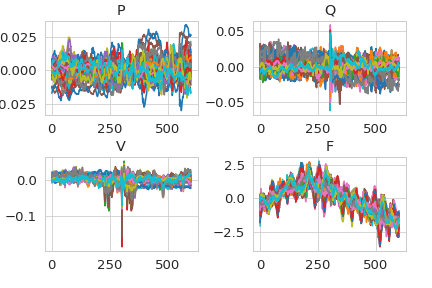}
    \end{subfigure}\hfill
    
    \caption{Samples of generated full event tensors designated as ``voltage events" caused by line trips. All time stamps on the horizontal axis correspond to $1/30$ seconds, and the full event window comprises $20$ seconds of synthetic event data. The presented data is scaled to per unit values and each timeseries is shifted to have a mean of zero.}
    \label{fig:vgensamples_trips}
\end{figure*}

\begin{figure*}[t]
    \centering
    
    \begin{subfigure}{0.33\linewidth}
        \includegraphics[width=\linewidth]{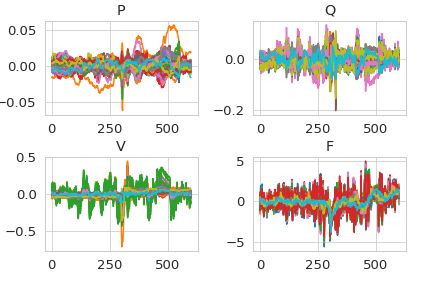}
    \end{subfigure}\hfill
    \begin{subfigure}{0.33\linewidth}
        \includegraphics[width=\linewidth]{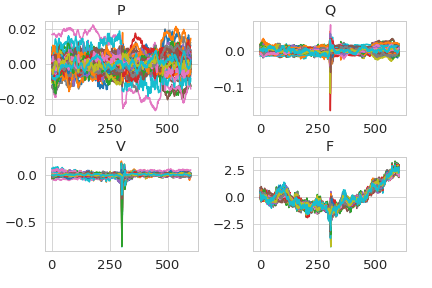}
    \end{subfigure}\hfill
    \begin{subfigure}{0.33\linewidth}
        \includegraphics[width=\linewidth]{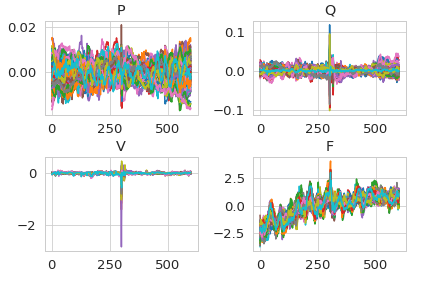}
    \end{subfigure}\hfill
    
    \caption{Samples of generated full event tensors designated as ``voltage events" caused by wind. All time stamps on the horizontal axis correspond to $1/30$ seconds, and the full event window comprises $20$ seconds of synthetic event data. The presented data is scaled to per unit values and each timeseries is shifted to have a mean of zero.}
    \label{fig:vgensamples_wind}
\end{figure*}

\begin{figure*}[t]
    \centering
    
    \begin{subfigure}{0.33\linewidth}
        \includegraphics[width=\linewidth]{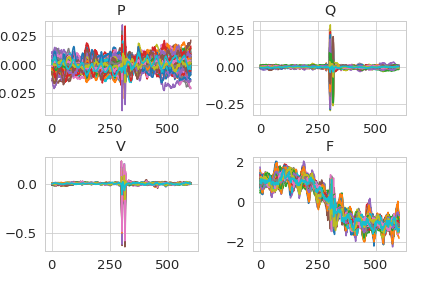}
    \end{subfigure}\hfill
    \begin{subfigure}{0.33\linewidth}
        \includegraphics[width=\linewidth]{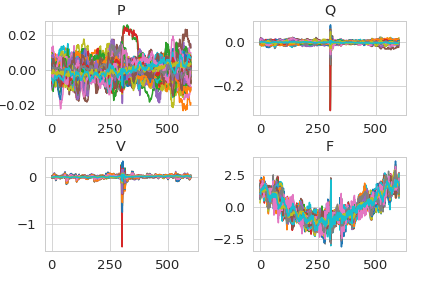}
    \end{subfigure}\hfill
    \begin{subfigure}{0.33\linewidth}
        \includegraphics[width=\linewidth]{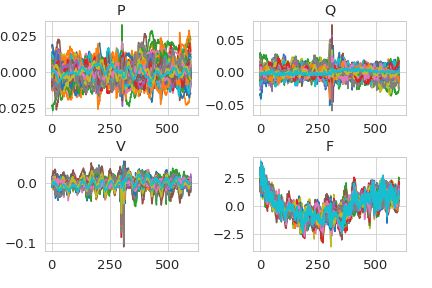}
    \end{subfigure}\hfill
    
    \caption{Samples of generated full event tensors designated as ``voltage events" caused by equipment failures. All time stamps on the horizontal axis correspond to $1/30$ seconds, and the full event window comprises $20$ seconds of synthetic event data. The presented data is scaled to per unit values and each timeseries is shifted to have a mean of zero.}
    \label{fig:vgensamples_equipment}
\end{figure*}

Sample images of Voltage Events in pmuBAGE are demonstrated in Figures \ref{fig:vgensamples_lightning} through \ref{fig:vgensamples_equipment}. Here, we have increased the granularity of our event labels into further groups of ``lightning strikes", ``line trips", ``wind", and ``equipment failures".

\begin{figure*}[t]
    \centering
    
    \begin{subfigure}{0.33\linewidth}
        \includegraphics[width=\linewidth]{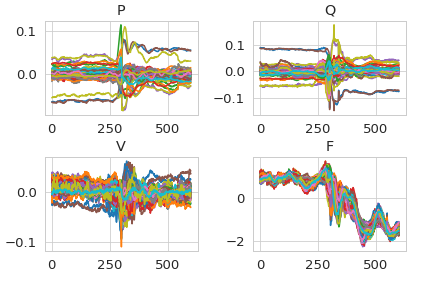}
    \end{subfigure}\hfill
    \begin{subfigure}{0.33\linewidth}
        \includegraphics[width=\linewidth]{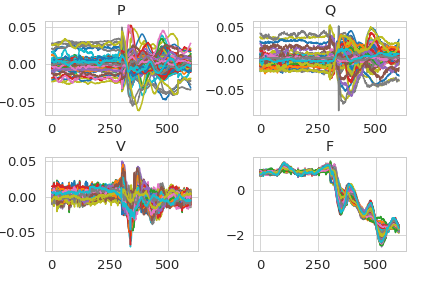}
    \end{subfigure}\hfill
    \begin{subfigure}{0.33\linewidth}
        \includegraphics[width=\linewidth]{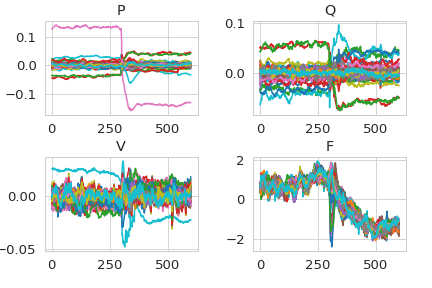}
    \end{subfigure}\hfill
    
    \begin{subfigure}{0.33\linewidth}
        \includegraphics[width=\linewidth]{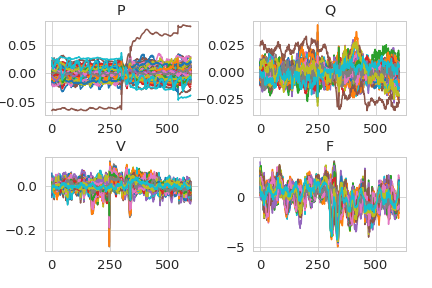}
    \end{subfigure}\hfill
    \begin{subfigure}{0.33\linewidth}
        \includegraphics[width=\linewidth]{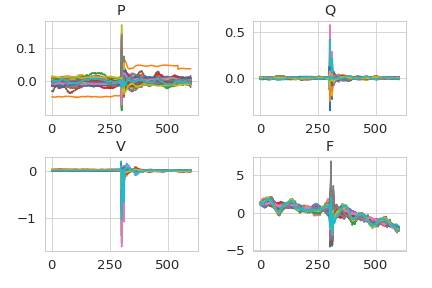}
    \end{subfigure}\hfill
    \begin{subfigure}{0.33\linewidth}
        \includegraphics[width=\linewidth]{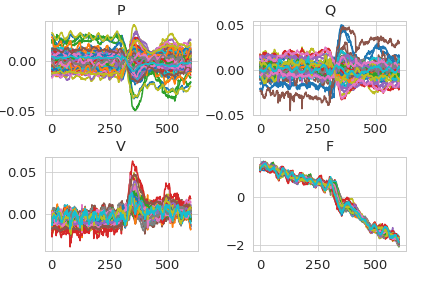}
    \end{subfigure}\hfill
    
    \caption{Samples of generated full event tensors designated as ``frequency events" caused by generator trips. All time stamps on the horizontal axis correspond to $1/30$ seconds, and the full event window comprises $20$ seconds of synthetic event data. The presented data is scaled to per unit values and each timeseries is shifted to have a mean of zero.}
    \label{fig:fgensamples_trip}
\end{figure*}

\begin{figure*}[t]
    \centering
    
    \begin{subfigure}{0.33\linewidth}
        \includegraphics[width=\linewidth]{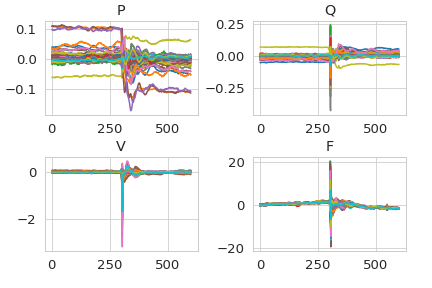}
    \end{subfigure}\hfill
    \begin{subfigure}{0.33\linewidth}
        \includegraphics[width=\linewidth]{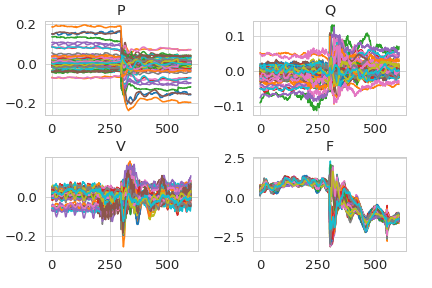}
    \end{subfigure}\hfill
    \begin{subfigure}{0.33\linewidth}
        \includegraphics[width=\linewidth]{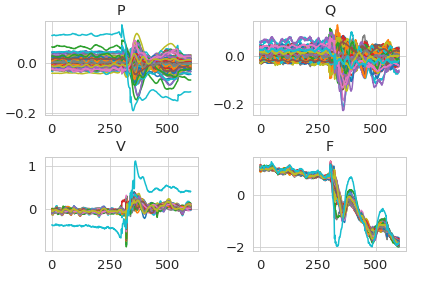}
    \end{subfigure}\hfill
    
    \begin{subfigure}{0.33\linewidth}
        \includegraphics[width=\linewidth]{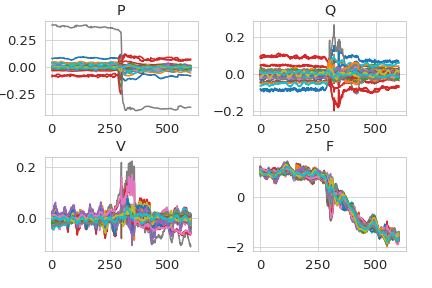}
    \end{subfigure}\hfill
    \begin{subfigure}{0.33\linewidth}
        \includegraphics[width=\linewidth]{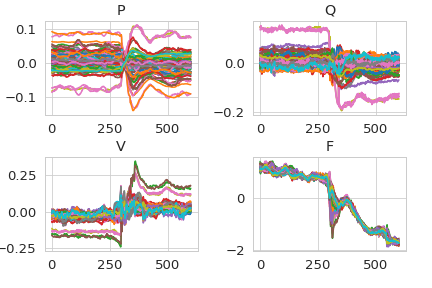}
    \end{subfigure}\hfill
    \begin{subfigure}{0.33\linewidth}
        \includegraphics[width=\linewidth]{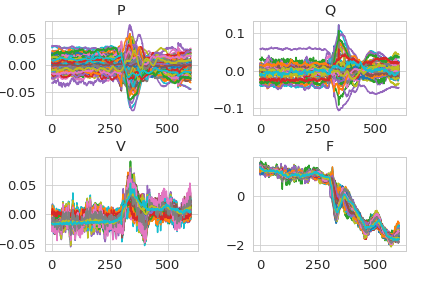}
    \end{subfigure}\hfill
    
    \caption{Samples of generated full event tensors designated as ``frequency events" caused by on-premises generator equipment failures. All time stamps on the horizontal axis correspond to $1/30$ seconds, and the full event window comprises $20$ seconds of synthetic event data. The presented data is scaled to per unit values and each timeseries is shifted to have a mean of zero.}
    \label{fig:fgensamples_equipment}
\end{figure*}

Sample images of Frequency Events in pmuBAGE are demonstrated in Figures \ref{fig:fgensamples_trip} and \ref{fig:fgensamples_equipment}. Here, we have increased the granularity of our event labels into further groups of ``generator trips", and ``on premises generator equipment failures".

Each color in these figures represents a different PMU, while each group of four figures comprises one event (where $P$ is real power data, $Q$ is reactive power data, $V$ is voltage magnitude data, and $F$ is frequency data).

\section{Conclusion}
The synthesized PMU dataset created in this work, pmuBAGE, is highly realistic to the human eye and does not significantly degrade important training evaluation metrics compared to the original measured PMU dataset from which it was trained. Thus, pmuBAGE may serve the scientific community as a standard benchmarking tool for algorithms which need PMU dynamic event data to train and validate on. 

Researchers should be aware of some pitfalls to this synthesized dataset. Four, in particular, are notable. First, there is a lack of realistic banding for frequency data - that is, the tendency for several PMUs to have the same frequency behavior, especially during a frequency event. Instead, as a direct result of the method used to train it, pmuBAGE tends to have PMUs act similarly in frequency data but with a slight spread in their exact values. Second, there is a small ``wrong direction" effect in voltage events voltage data. Actual voltage data undergoing a voltage event will primarily have PMUs drop their voltage values, while pmuBAGE will have a small amount of PMUs raise their values very slightly. Third, pmuBAGE occasionally displays Inter-Area Oscillations that do not dampen as quickly as actual data; however, the difference is relatively small. Finally, the generated events are a little more similar to each other than one would expect from a real dataset. 

While we wish to eliminate these problems in future versions of this dataset, we believe that the current incarnation is sufficiently realistic to serve its purpose to the community - as a training, validating, and benchmarking dataset. We encourage any researcher to utilize this synthetic PMU dataset for these purposes.

\section{Acknowledgements}
We sincerely thank James Follum at the Pacific Northwest National Laboratory for reviewing the generated event dataset in detail.

\printbibliography



\end{document}